\documentclass{article}

\usepackage{PRIMEarxiv}

\usepackage[utf8]{inputenc} %
\usepackage[T1]{fontenc}    %
\usepackage{hyperref}       %
\usepackage{url}            %
\usepackage{booktabs}       %
\usepackage{amsfonts}       %
\usepackage{nicefrac}       %
\usepackage{microtype}      %
\usepackage{lipsum}
\usepackage{fancyhdr}       %
\usepackage{graphicx}       %
\usepackage{natbib}
\usepackage[most]{tcolorbox}
\usepackage{multirow}
\usepackage{threeparttable}
\usepackage{listings}
\usepackage{listings}
\usepackage[most]{tcolorbox}
\usepackage{listings}
\usepackage{xcolor} %
\graphicspath{{media/}}     %

\title{Physics Supernova: AI Agent Matches Elite Gold Medalists at IPhO 2025}

\begin{document}

\maketitle
\vspace{-7em}
\begin{center}
Jiahao Qiu$^{*1}$, Jingzhe Shi$^{*\#2}$, Xinzhe Juan$^{3,4}$, Zelin Zhao$^5$, Jiayi Geng$^{1}$, Shilong Liu$^{1}$, Hongru Wang$^{1}$, Sanfeng Wu$^6$,  Mengdi Wang$^1$
\end{center}

\begin{center}
\textsuperscript{1}AI Lab, Princeton University \quad
\textsuperscript{2}IIIS, Tsinghua University \quad
\textsuperscript{3}Shanghai Jiao Tong University \quad \\
\textsuperscript{4}University of Michigan \quad
\textsuperscript{5}King's College London\quad
\textsuperscript{6}Department of Physics, Princeton University \quad

\end{center}

\vspace{3em}

\begingroup
 \let\thefootnote\relax
\footnotetext{$^*$ These authors contributed equally to this work.}
\footnotetext{$^\#$ IPhO 2021 gold medalist (ranked \#10), IPhO 2022 marker.}
\endgroup

\begin{abstract}
Physics provides fundamental laws that describe and predict the natural world.
AI systems aspiring toward more general, real-world intelligence must therefore demonstrate strong physics problem solving abilities: to formulate and apply physical laws for explaining and predicting physical processes.
The International Physics Olympiad (IPhO)--the world's most prestigious physics competition--offers a rigorous benchmark for this purpose. We introduce Physics Supernova, an AI agent system with superior physics problem-solving abilities that match elite IPhO gold medalists. In IPhO 2025 theory problems, Physics Supernova attains 23.5/30 points, ranking 14\textsuperscript{th} of 406 contestants and surpassing the median performance of human gold medalists. We extensively analyzed Physics Supernova's capabilities and flexibility across diverse physics tasks. These results show that principled tool integration within agent systems can deliver competitive improvements in solving challenging science problems. The codes are available at \href{https://github.com/CharlesQ9/Physics-Supernova}{https://github.com/CharlesQ9/Physics-Supernova}.

\end{abstract}

\newenvironment{mdblock}%
  {\markdownSetup{underscores=false, codeSpans=false, hashEnumerators=false, fencedCode=false}\begin{markdown}}%
  {\end{markdown}}
\newcommand{\problemname}{Problem}
\newcommand{\answername}{GT Answer}
\newcommand{\wowolf}{\textbf{w.o.wolfTool Answer:}}
\newcommand{\wwolf}{\textbf{w.wolfTool Answer:}}
\newcommand{\somespace}{\hspace{12pt}}
\newtcolorbox{qaBlock}{
  enhanced,
  colback=white,
  colframe=black!15,
  boxrule=0.4pt,
  arc=2mm,
  left=2mm,right=2mm,top=1.0mm,bottom=1.0mm,
  before skip=8pt, after skip=8pt
}

\newcommand{\AnswerAligned}[4]{%
  \noindent
  \begin{tabular*}{\linewidth}{@{}l@{\extracolsep{\fill}}c@{}r@{}}%
    \mbox{\textbf{\answername#1}\ #2} &
    \mbox{\wowolf\ #3} &
    \mbox{\wwolf\ #4} \\
  \end{tabular*}%
}

\lstdefinestyle{mystyle}{
  basicstyle=\ttfamily\small,
  keywordstyle=\color{blue},
  stringstyle=\color{red},
  commentstyle=\color{green!50!black},
  showstringspaces=false,
  frame=single,
  breaklines=true,
  rulecolor=\color{black}, %
}

\lstnewenvironment{python}{\lstset{style=mystyle,language=Python}}{}

\definecolor{shallowblue}{RGB}{35,75,237} %
\colorlet{shallowerblue}{shallowblue!60!white}
\definecolor{wolframred}{RGB}{191,10,48}    %

\newcommand{\None}{\textcolor{gray}{None}}
\newcommand{\Low}{\textcolor{gray}{Low}}
\newcommand{\Med}{\textcolor{shallowblue}{Med}}
\newcommand{\High}{\textcolor{wolframred}{High}}
\newcommand{\Common}{\textcolor{gray}{Common}}
\newcommand{\Rare}{\textcolor{shallowblue}{Rare}}
\newcommand{\Novel}{\textcolor{wolframred}{Novel}}
\newcommand{\NoImage}{\textcolor{gray}{No Image}}
\newcommand{\Understand}{\textcolor{shallowerblue}{Understand}}
\newcommand{\RoughMeasure}{\textcolor{shallowblue}{Rough Meas}}
\newcommand{\AccurateMeasure}{\textcolor{wolframred}{Accurate Meas}}

\section{Introduction}
\label{sec:intro}

\begin{quote}\itshape
“The supreme task of the physicist is to arrive at those universal elementary laws from which the cosmos can be built up by pure deduction.”
\end{quote}
\begin{flushright}
—\,\textit{Albert Einstein} %
\end{flushright}

Physics aims to formulate the fundamental laws that govern the behavior of the universe, from the dynamics of subatomic particles and macroscopic objects to the evolution of cosmic structures on the largest scales~\cite{feynman1965character,peebles1993principles}. Expressed through the compact formalism of mathematics and physical theory, these laws offer the most concise representations of the complex physical world and enable reliable predictions of future events~\cite{wigner1960unreasonable}. For both humans and AI systems, mastery of physics entails constructing rigorous abstractions, such as well-defined state variables, conserved quantities, and causal structures, that underpin the explanation, prediction, and control of physical systems~\cite{noether1918invariante,astrom2008feedback}. As AI systems increasingly integrate into the physical world and advance toward Artificial General Intelligence (AGI) and potentially Artificial Super intelligence (ASI), a deep grounding in physics, along with the ability to solve physics problems through its abstractions, becomes a critical foundation for developing competent and reliable intelligence~\cite{lee2008cps,bostrom2014superintelligence,lecun2022path}. 
In this work, we investigate how to enhance AI systems' physics problem-solving capabilities using agent-based architectures. To assess progress in this challenging domain, we benchmark our models on the theory problems of the 2025 International Physics Olympiad\footnote{IPhO 2025 official website: \url{https://www.ipho2025.fr/}.} (IPhO~\cite{ipho})
, a globally recognized competition that emphasizes deep conceptual understanding, abstraction, and advanced problem-solving in physics.

\textbf{The International Physics Olympiad}, or IPhO~\cite{ipho}, widely viewed as a prestigious physics competition, stands alongside the International Mathematical Olympiad (IMO~\cite{imo}) in reputation and impact. The 2025 International Physics Olympiad (IPhO), held in July in France~\cite{ipho2025}, featured three theory problems designed to test contestants’ conceptual understanding, logical reasoning, and problem-solving skills in physics. Unlike earlier benchmarks evaluating physics problem-solving abilities~\cite{xiang2025seephysdoesseeinghelp,he2024olympiadbenchchallengingbenchmarkpromoting,qiu2025phybenchholisticevaluationphysical} that might risk data contamination, use coarse evaluation (e.g., final-answer-only scoring), offer limited novelty, and lack assessments of precise figure reading/measurement, etc., IPhO 2025 Theory Problems serve as a rigorous benchmark for evaluating an AI system’s mastery of physics. These problems exhibit several distinctive features: (1) they were newly released in July 2025~\cite{ipho2025}; (2) they employ fine-grained, part-level scoring, enabling detailed assessment of reasoning and solution steps; (3) they incorporate uncommon physics models that challenge standard approaches; and (4) they include explicit figure-based measurement requirements (see Table ~\ref{tab:theory problem description}), demanding precise interpretation of visual data. In this work, we utilize \textbf{IPhO 2025 Theory Problems} as the benchmark for evaluating AI's capability in physics.

IPhO contestants are provided with external tools, including calculators and fundamental constant tables; moreover, expert physicists routinely work with extra tools:
these facts highlight the importance of external tool use in solving physics problems and for physicists in general. However, previous work on employing AI systems for solving physics problems mainly focuses on the performance of base Large Language Models (LLMs), or only with simple best-of-N style test-time scaling methods~\cite{qiu2025phybenchholisticevaluationphysical,snell2024scalingllmtesttimecompute,testtimescaling}. In this body of work, LLMs are not equipped with any extra tools: the reasoning, problem-solving and physics knowledge of the LLMs are tested, mainly aiming at benchmarking and improving plain LLMs. However, they cannot represent the physics problem-solving ability of state-of-the-art AI architectures, which are usually composed of LLM agents and tools.

\textbf{LLM-based agent systems} demonstrate significant advantages over standalone LLM in planning, generalization, and complex reasoning. They have been used to perform complicated tasks with little human interaction~\cite{wang2025surveyevolutionlanguagemodelbased,llmzeroshot}. For example, ReAct~\cite{yao2023react} introduces the Reasoning-Acting loop for agents. More recently, the idea of a self-evolving agent has been proposed, with even fewer designs based on humans~\cite{gao2025surveyselfevolvingagentspath, qiu2025alitageneralistagentenabling, feng2025groupingrouppolicyoptimizationllm, agentdistill}. However, this line of work mainly examines general-purpose agents with a focus on possible virtual assisting style tasks (e.g., GAIA~\cite{mialon2023gaia}) or other domains like math problems~\cite{geminideepthinkimo2025}, History~\cite{histagent}, etc., not covering the domain of complex physics problems. 

In this work, with a focus on physics reasoning and problem-solving ability, we introduce \textbf{Physics Supernova}, an agent system equipped with physics-oriented tools targeting physics problem-solving. Physics Supernova equips LLMs with tools to improve their reasoning about the physical world. 
For understanding schematic diagrams and accurately extracting data from figures, we add the Image Analyzer Tool; for self-review and refinement, we provide the Answer Reviewer Tool. With these tools, Physics Supernova achieves \textbf{Gold Medal} in IPhO 2025 Theory Problems: it ranks top $10\%$ on all three problems tested, and ranks 14\textsuperscript{th} among 406 human contestants in IPhO 2025 Theory Problems. Moreover, we further explore the possibility to improve Physics Supernova's problem-solving skill with specialized Physics QA tools (e.g.WolframAlpha~\cite{WolframAlpha}). To conclude, we summarize our main contribution as follows:
\begin{itemize}
    \item [\textbf{1.}] We develop Physics Supernova, an agent system combining Large Language Models with physics-specific tools, exhibiting strong problem-solving capabilities across a wide range of tasks.
    \item [\textbf{2.}] We show that Physics Supernova achieves gold-medalist-level performance on IPhO 2025 Theory Problems, ranking 14\textsuperscript{th} among all 406 human contestants worldwide, exceeding the official gold medalist median score.
    \item [\textbf{3.}] We conduct further analysis and experiments to show Physics Supernova's capability and flexibility for physics problem solving tasks.
\end{itemize}

\begin{figure}[!t]
  \centering
  \includegraphics[width=\linewidth]{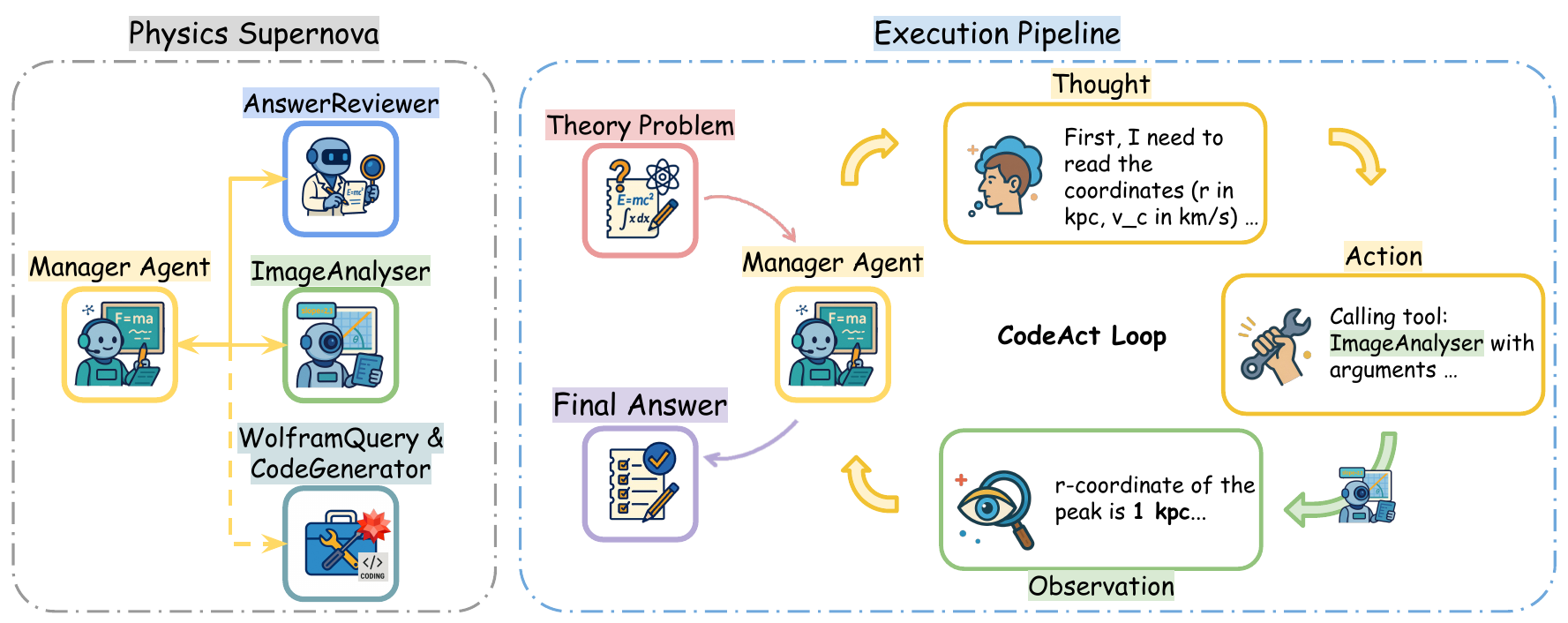}
  \caption{Our proposed agent system: Physics Supernova, for solving theory problems in physics.}

  \label{fig:agent pipeline}
\end{figure}

\begin{table}[!t]
\centering
\resizebox{0.99\columnwidth}{!}{%
\begin{tabular}{c c l c c c c}
\toprule
\multirow{2}{*}{\textbf{Problem}} &
\multirow{2}{*}{\textbf{Subpart}} &
\multicolumn{1}{l}{\textbf{Subpart Content}} &
\multicolumn{1}{c}{\textbf{Model}} &
\multicolumn{1}{c}{\textbf{Computation}} &
\multicolumn{1}{c}{\textbf{Image}} &
\multicolumn{1}{c}{\textbf{Top}} \\
 & & \multicolumn{1}{l}{\textbf{\{\# Scoring Points$^\dagger$\}}} &
\multicolumn{1}{c}{\textbf{Novelty}} &
\multicolumn{1}{c}{\textbf{Difficulty}} &
\multicolumn{1}{c}{\textbf{Ability}} &
\multicolumn{1}{c}{\textbf{10\% Score}} \\
\midrule
\multirow{4}{*}{\parbox{3.0cm}{\centering Theory Problem 1\\(Modern Physics)}} 
  & Part~A & Bohr Model~\{18\} & \Common & \Low & \NoImage & \multirow{4}{*}{9.0/10.0} \\
  & Part~B & Rotation of a Galaxy~\{18\} & \Rare & \Med & \AccurateMeasure \\
  & Part~C & Mass Distribution of a Galaxy~\{24\} & \Rare & \Med & \AccurateMeasure \\
  & Part~D & Tully-Fisher Relation \& MOND~\{23\} & \Novel & \Med & \RoughMeasure \\
\midrule
\multirow{3}{*}{\parbox{3.0cm}{\centering Theory Problem 2\\(Thermodynamics \& Dynamics)}} 
  & Part~A & Pulling on a Submerged Tube~\{9\} & \Rare & \Med & \Understand & \multirow{3}{*}{5.0/10.0} \\
  & Part~B & Two-Part Barometric Tube~\{10\} & \Rare & \High & \Understand \\
  & Part~C & Cox's Timepiece~\{38\} & \Novel & \High & \Understand \\
\midrule
\multirow{3}{*}{\parbox{3.0cm}{\centering Theory Problem 3\\(Thermodynamics \& Dynamics)}} 
  & Part~A & Nucleation \& Growth of Bubbles~\{26\} & \Rare & \Med & \RoughMeasure & \multirow{3}{*}{8.0/10.0} \\
  & Part~B & Acoustic Emission of Bubbles~\{20\} & \Rare & \Med & \Understand \\
  & Part~C & Popping Champagne~\{19\} & \Rare & \Med & \Understand \\
\bottomrule
\end{tabular}}
\vspace{3mm}
\caption{Theory problems of IPhO 2025~\cite{ipho2025}: Subpart contents (and number of scoring points$^\dagger$ for each subpart, obtained from official solutions at \url{https://ipho.olimpicos.net/}), physics model novelty (among Physics Olympaids), computation difficulty levels, required image-reading skills, and top $10\%$ human scores. For physics model novelty: \Common~means such a model appears often in Physics Olympiads; \Rare~means such a physics model rarely appears in previous Physics Olympiads; while \Novel~represents physics models with high novelty, which almost originally appeared in Physics Olympiads. For computation Difficulty: \Low~means easy to compute, without tedious details and easy to obtain a correct result; \Med~means the problem requires complicated computation to a certain extent; while \High~means requiring tedious computation and careful attention for applying formulas, lacking of which could potentially lose a lot of points. For Image reading requirements: \NoImage~ means no image, \Understand~ represents cases where contestants need to understand the image (but measurements are not required), while \RoughMeasure~requires a non-accurate rough measurement from the provided image, and \AccurateMeasure~requires accurate reading from the image. Examples of these problems are shown in Appendix \ref{app: score criteria example}.}

\label{tab:theory problem description}
\end{table}

\section{Method}

\subsection{Agent Architecture}
We introduce \textbf{Physics Supernova}, an AI agent system designed to solve complex physics theory problems. It follows the \texttt{CodeAgent} architecture from the \texttt{smolagents} framework~\cite{smolagents}. As illustrated in Figure~\ref{fig:agent pipeline}, the system consists of a \textbf{Manager Agent}~$\mathcal{M}$ and a set of domain-specific tools~$\mathcal{D} = \{d_i\}_{i=1}^{n}$, where each $d_i$ is physics-problem-oriented tool.

Unlike prior work in mathematical problem-solving that often relies on fixed, manually hard-coded workflows~\cite{huang2025gemini25procapable, lin2025leanstarlearninginterleavethinking}, our approach emphasizes \textbf{flexible self-planning}. Inspired by the design philosophy of \textit{ minimum pre-definition and maximal autonomy}~\cite{qiu2025alitageneralistagentenabling}, the Manager Agent is granted access to the tool set $\mathcal{D}$ but is not provided with a hard-coded predefined execution graph or action script. The agent is able to call different Tools according to its progress made in its process of solving problems.

When presented with a physics theory problem $\mathcal{Q} = \{(q_j, s_j)\}_{j=1}^{m}$—where each $q_j$ is a sub-question and $s_j$ contains the associated visual data: the agent solves problems gradually, constantly updating its trajectory~$\mathcal F$ of known information, assumptions, and relevant physical context.

The agent then operates in an iterative \textbf{Reason--Act} loop. In each round, it generates a natural language reasoning step to describe the current objective and justify the selection of a tool $d_i$ (\textit{Reason}); it then produces code for calling tools; the tools are called to produce an intermediate observation $o_t$ (\textit{Act}). This observation is incorporated into the next reasoning step, allowing the agent to refine its understanding of the problem. The loop continues until the final answers are produced for all subquestions in $\mathcal{Q}$.

\subsection{Physics-Problem-Oriented Toolset}

Here, we discuss the physics-problems-oriented tools we equip the manager agent with: the \textit{ImageAnalyzer} and the \textit{AnswerReviewer}. 

\textbf{\textit{ImageAnalyzer}}:
Reading experimental results and extracting critical information from figures is important for expert physicists, and is critical to some Physics Olympiad Problems. However, current LLMs show limited performance in providing accurate measurements on visual data such as data figures, images, and schematic representations. To enhance image handling, \textbf{\textit{ImageAnalyzer}} routes a high-res image to a dedicated Vision Language Model, addressing precise tasks such as reading numeric values and making measurements. We will discuss more about how this improves the accuracy of the information extracted from the visual data in Section \ref{sec: image analyzer}. The prompts for the Image Analyzer are available in the Appendix \ref{app: image analyzer tool prompt}.

\textbf{\textit{AnswerReviewer}}:
Physicists routinely evaluate whether their theoretical results are physically meaningful. This involves analyzing whether the outcomes exhibit reasonable physical properties or align with established principles, essentially, whether they make sense within known constraints. Such scrutiny is key to assessing solutions and sometimes led to major breakthroughs. For example, in the famous example of the "ultraviolet catastrophe"~\cite{RayleighJeans_UltravioletCatastrophe}, the prediction from classical physics that black-body radiation should diverge at short wavelengths was unphysical and was not supported by experimental data. This paradox prompted Max Planck ~\cite{Planck1900} to introduce his formulation of black-body radiation, laying the foundation for quantum mechanics. To enhance AI's ability of rethinking, \textbf{\textit{AnswerReviewer}} is provided: it classifies likely error types and locates erroneous expressions through the process. We provide ablations studies where the reviewing tool improves performance in Section \ref{sec: ablation study}. The prompts for the Answer Reviewer are available in Appendix \ref{app: answer reviewer tool prompt}.

With only \textbf{\textit{ImageAnalyzer}} and \textbf{\textit{AnswerReviewer}}, Physics Supernova empowered by state-of-the-art LLM (Gemini 2.5 Pro~\cite{comanici2025gemini25pushingfrontier}) can achieve medium gold-level performance in IPhO 2025 Theory Problems (Section \ref{sec: experiment}). Moreover, this system supports the integration of additional advanced physics-related tools, such as the WolframAlpha question-answering engine (Section \ref{sec: wolframalpha}) that can assist with computationally intensive physics tasks.

\section{Experiment: Physics Supernova Excels in IPhO 2025 Theory Problems}
\label{sec: experiment}

\subsection{Experiment Setup}

\paragraph{Benchmarking Dataset}

IPhO 2025 has 3 theory problems and 2 experimental problems in which each problem counts for $10.0$ points, adding up to $50.0$ points in total. Among $406$ contestants from more than $90$ countries and regions, $37$ ones with the highest total scores are awarded the Gold Medal. Theory score is the sum of all three theory problems. Detailed contents are shown in Table \ref{tab:theory problem description}. When ranked by difficulty, T2 is the most difficult one with the highest human score $10\%$ of $5.0/10.0$; T3 is easier with the highest human score $10\%$ of $8.0/10.0$; human competitors achieve the highest score in T1, with a top score $10\%$ of $9.0/10.0$. The minimum and median theory scores for gold medalists are $19.4$ \& $22.8$. When ranked by theory score only, the min \& median theory score for top $37$ contestants are $20.8$ \& $23.0$.

\paragraph{Methods}

We test Physics Supernova on each of the three theory problems, with chosen pre-defined tools. As smolagents~\cite{smolagents} lack a built-in summarization memory, we implemented a lightweight summarizing tool to summarize existing progress, which is provided beside the \textbf{\textit{ImageAnalyzer}} and \textbf{\textit{AnswerReviewer}} tools. Throughout the experiments, we use Gemini 2.5 Pro~\cite{comanici2025gemini25pushingfrontier} for the agent system and all the LLM-requiring tools.

\paragraph{Metrics}

For each theory problem, as shown in Table \ref{tab:theory problem description}, there are $3$ or $4$ major parts (Parts A, B, C, and D), where each part consists of multiple scoring points. The score for some certain part is obtained by summing up the scored points addressed in that particular part. To be more specific, for problem $\mathcal{Q} = \{(q_j, s_j)\}_{j=1}^{m}$, its score $\mathcal S=\sum_i\mathcal S_i$ where $\mathcal S_i$ is the score of each part, obtained by summing-up the score of each scoring points, $P_{ij}$ with score $S_{ij}$:
$$
\mathcal S_i = \sum_{j}S_{ij}*\mathbf 1_{\left[\text{$P_{ij}$ is addressed}\right]}\text{, where $S_i$ represents score for Part $i$.}
$$

The official scoring criteria are obtained online\footnote{We obtain official solutions and scoring criteria from \url{https://ipho.olimpicos.net/}.}, examples of which are provided in Appendix \ref{app: score criteria example}. After the LLM answers are generated, human experts\footnote{Our human experts include previous Olympiad medalists.} then score the answers in detail according to the official scoring criteria.

For each problem, the experiment is carried out $5$ times, where mean and standard deviation are reported in Table\ref{tab:agent_comparison}.

\subsection{Main Experimental Results}

\begin{table}[!t]
\centering
\resizebox{0.95\columnwidth}{!}{%
\begin{tabular}{cccccc}
\toprule
\textbf{Problem} & \textbf{Part A} & \textbf{Part B} & \textbf{Part C} & \textbf{Part D} & \textbf{SUM} \\
\midrule
\textbf{Theory 1} total score &$2.2$ &$2.5$ &$3.0$&$2.3$&$10.0$ \\
LLM Only & $2.20\pm0.00$ & $2.28\pm0.04$ & $2.06\pm0.13$ & $1.92\pm0.11$ & $8.46\pm0.16$ \\
\textbf{Physics Supernova} & $2.20\pm0.00$ & $2.20\pm0.00$ & $2.46\pm0.05$ & $2.16\pm0.05$ & $\boldsymbol{9.02 \pm 0.11}$ \\
Human Top $10\%$ &/&/&/&/&$\sim9.0$\\
\midrule
\textbf{Theory 2} total score &$1.3$&$2.0$&$6.7$& / &$10.0$ \\
LLM Only & $1.16 \pm 0.31$ & $1.08\pm0.19$ & $3.04\pm 1.02$ & / & $5.30\pm0.99$ \\
\textbf{Physics Supernova} & $1.30 \pm 0.00$ & $1.16 \pm 0.22$ & $3.62 \pm 0.79$ & / & $\boldsymbol{6.08 \pm 0.77}$ \\
Human Top $10\%$ &/&/&/&/&$\sim5.0$\\
\midrule
\textbf{Theory 3} total score &$4.3$&$3.3$&$2.4$ & / &$10.0$ \\
LLM Only & $3.70\pm0.07$ & $2.74\pm0.36$ & $1.22\pm0.18$ & / & $7.66\pm0.51$ \\
\textbf{Physics Supernova} & $4.02\pm0.22$ & $3.26\pm0.09$ & $1.12\pm0.11$ & / & $\boldsymbol{8.40\pm0.27}$ \\
Human Top $10\%$ &/&/&/&/&$\sim8.0$\\
\midrule
\textbf{Theory Part} total score &/& / &/ & / &$30.0$ \\
LLM Only & / & / & / & / & $21.4\pm1.1$ \\
\textbf{Physics Supernova} & / & / & / & / & $\boldsymbol{23.5 \pm 0.8}$ \\
Gold Medalists (mediam) &/&/&/&/&$22.8$\\
\bottomrule
\end{tabular}}
\vspace{1mm}
\caption{Experiment results for Physics Supernova (mean$\pm$std) across multiple problems and parts (with Gemini 2.5 Pro), for $5$ rounds. Our agent results rank top $10\%$ among human contestants on all three Theory problems.}
\label{tab:agent_comparison}
\end{table}

The results of our main experiments are shown in Table \ref{tab:agent_comparison}. The mean theory score of Gemini 2.5 Pro alone ranks 30\textsuperscript{th} among 406 contestants, while the mean theory score of \textbf{Physics Supernova ranked 14\textsuperscript{th} among all 406 contestants, surpassing the medium theory score of the gold medalists}. Noticeable, Physics Supernova ranks top $10\%$ on all three Theory problems tested. Comparing Table \ref{tab:theory problem description} and Table \ref{tab:agent_comparison}, we make the following observations.

\paragraph{Larger advantage for Physics Supernova on more difficult problems.}

Theory Problem 2 is the most difficult, Theory Problem 3 is easier, while Theory Problem 1 is the easiest among all three theory problems, as shown in the difficulty levels and human scores in Table \ref{tab:theory problem description}. For harder problems, Physics Supernova: (1) shows more obvious advantage compared to Human Top $10\%$ scores; and (2) shows larger advantage compared to using LLM Only.

\paragraph{Larger variance for LLM-based systems on more difficult problems.}

LLM results on these problems show a larger variance. In detail, LLM Only and Physics Supernova show larger variance (as reflected by SD) on Theory Problem 2, which is the most difficult problem. In contrast, for Theory Problem 1, which is the easiest problem among all three, the performance of AI-based systems shows low variance on these problems.

These results indicate that, with agent systems and dedicated tools
, current state-of-the-art LLMs have the ability to match elite gold medalists on IPhO level physics problems.

\section{Analysis}
\label{sec: impact study}
\subsection{Ablation Study}
\label{sec: ablation study}

To study the impact of each tool on the final scores, we replicate the protocol described in Section \ref{sec: experiment}, providing different tools for the agent system. We consider four settings: (1) Physics Supernova; (2) without \textit{ImageAnalyzer} tool; (3) without \textit{AnswerReviewer} tool; and (4) LLM only. For each theory problem, we perform $5$ independent runs and report the mean and standard deviation. The agents are powered by Gemini 2.5 Pro. The results are shown in Table \ref{tab:ablation study}. Comparing Table \ref{tab:theory problem description} and Table\ref{tab:ablation study}, we find:
\paragraph{\textit{ImageAnalyzer} helps with tasks requiring Accurate Image Measurements.} For example, as shown in Table \ref{tab:theory problem description}, the theory problem requires accurate measurements from figures, and in this case, the Image Analyzer successfully improves the performance of the model. This indicates that delegating high-accuracy image analysis to \textit{ImageAnalyzer} improves the overall score. We present a case study in Section \ref{sec: image analyzer} for better illustration.

\paragraph{\textit{AnswerReviewer} raises overall scores via post-hoc review.} In most problems (especially non-easy ones), removing the Answer Reviewer reduces performance. For example, we see a performance drop for overall scores for all three theory problems after removing the Answer Reviewer. This implies that, equipped with \textit{AnswerReviewer} for reviewing, locating errors and providing feedback, Physics Supernova can improve performance for many cases.

\begin{table}[!t]
\centering
\resizebox{0.95\columnwidth}{!}{%
\begin{tabular}{lcccc}
\toprule
\textbf{Method} & \textbf{Theory 1} & \textbf{Theory 2} & \textbf{Theory 3} & \textbf{Theory Part} \\
\midrule
Total score & $10.0$ & $10.0$ & $10.0$ & $30.0$ \\
\midrule
\textbf{Physics Supernova} & $\boldsymbol{9.02\pm0.11}$ & $\boldsymbol{6.08\pm0.77}$ & $\boldsymbol{8.40\pm0.27}$ & $\boldsymbol{23.5\pm0.8}$ \\
w.o.ImgTool & $8.58\pm0.19$ & $5.98\pm0.54$ & $8.26\pm0.24$ & $22.8\pm0.6$ \\
w.o.RevTool & $8.62\pm0.16$ & $5.54\pm0.61$ & $8.26\pm0.15$ & $22.4\pm0.6$ \\
LLM Only & $8.46\pm0.16$ & $5.30\pm0.99$ & $7.66\pm0.51$ & $21.4\pm1.1$ \\
\bottomrule
\end{tabular}}
\vspace{1mm}
\caption{Ablation study results for four experiment settings: Physics Supernova, Physics Supernova without \textbf{\textit{AnswerReviewer}} (RevTool), Physics Supernova without \textbf{\textit{ImageAnalyzer}} (ImgTool), and using LLM only.}
\label{tab:ablation study}
\end{table}

\subsection{Case Study: Image Analyzer Tool}
\label{sec: image analyzer}

\begin{figure}[!t]
  \centering
  \includegraphics[width=\linewidth]{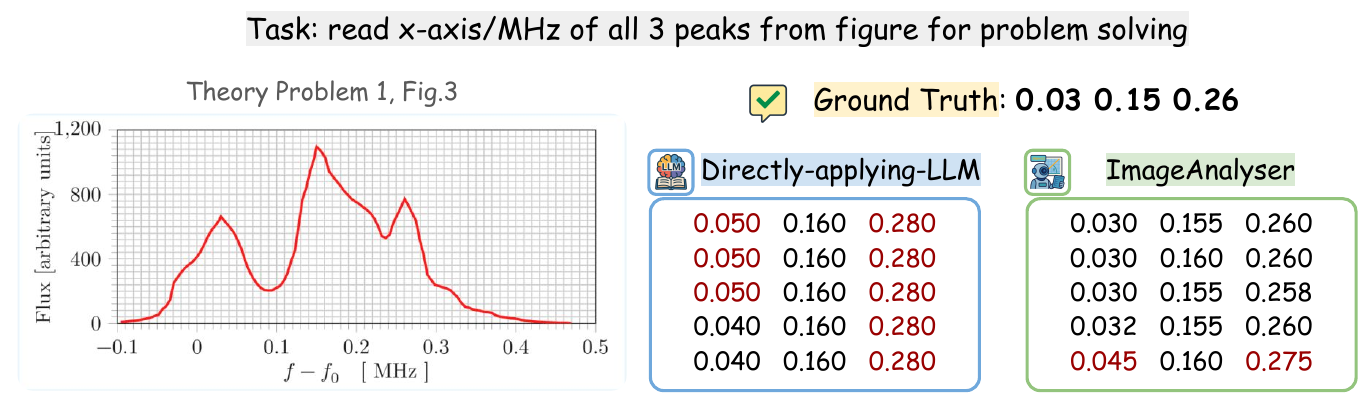}
  \caption{Effect of \textit{ImageAnalyzer} on Theory Problem 1 Part C. The problem here requires accurate measurement from a figure (shown left). We show $5$ repeated experiments of directly applying LLMs and using Image Analyzer Tool (shown right), with reading difference $> 0.01 \text{MHZ}$ colored in red. As shown, the improvement mainly comes from a reduction in measurement error.}
  \label{fig:ExampleImg to be read}
\end{figure}

We zoom in on Theory Problem 1 Part C to see how Image Analyzer Tool helps improve scores. In particular, this problem requires the contestants to accurately read a figure to solve the problem. The specific task related to the image is shown in Figure \ref{fig:ExampleImg to be read}.

As shown in Figure \ref{fig:ExampleImg to be read}, using a specialized Image Reading LLM as Image Analyzer Tool excels directly exposing the manager agent to all images of a single problem, which reduces the mean absolute error (MAE) from $0.015$ to $0.004$. This example further shows that, by using Image Analyzer Tool, Physics Supernova improves its ability to accurately read from figures, thus improving its overall performance.

\subsection{Domain-Specific Tools: WolframAlpha QA Tool}
\label{sec: wolframalpha}

\begin{figure}[!t]
  \centering
  \includegraphics[width=\linewidth]{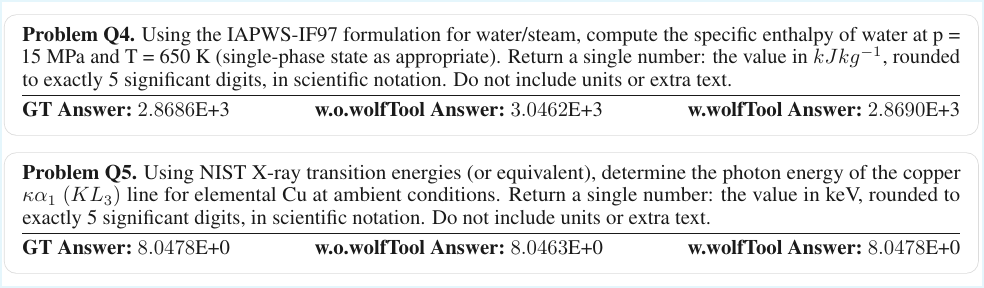}
  \caption{Examples of 10 expert-knowledge-requiring problems we generate, listed in Appendix \ref{app: wolfalpharesults}. These examples require expert domain knowledge that can only be obtained through web search/expert database queries.}
  \label{fig:Example of wolframalpha problem}
\end{figure}

Physics research usually requires expert domain knowledge. In Physics Olympiads, some reference information (e.g., physical constants, specific physics formulas, etc.) is provided as information for contestants to use: this is not the case for more complicated Physics researches. In these expert-level cases, accessing domain knowledge is very important. Figure \ref{fig:Example of wolframalpha problem} provides examples of this case.

To better handle domain-specific queries in Physics Supernova, we further equip Physics Supernova with a question-answering (QA) tool for expert domain knowledge. To be specific, we utilize WolframAlpha~\cite{WolframAlpha}: a computational knowledge engine capable of providing accurate and concise results for science-related queries.

In order to study Physics Supernova's performance on more expert tasks with WolframAlpha QA Tool, we generate $10$ problems which require expert knowledge (most efficiently obtained from standard references or WolframAlpha), listed in Appendix \ref{app: wolfalpharesults}. As shown in Figure \ref{fig:Example of wolframalpha problem} are examples of these problems. Table~\ref{tab: wolfalpharesults} reports the performance with and without the WolframAlpha tool. The result shows that WolframAlpha Tool improves the ability of Physics Supernova to solve problems that require expert physics knowledge.

\begin{table}[!t]
\centering
{\setlength{\tabcolsep}{18pt} %
\resizebox{0.95\columnwidth}{!}{
\begin{tabular}{lcc}
\toprule
 & \textbf{w.o. WolframAlpha Tool} & \textbf{w. WolframAlpha Tool} \\
\midrule
\textbf{$\#$ 3-digit Accurate Answers} & 3/10 & 9/10 \\
\textbf{$\#$ 4-digit Accurate Answers} & 2/10 & 6/10 \\
\bottomrule
\end{tabular}}%
}
\vspace{1mm}
\caption{Number of accurate answers for expert-knowledge-requiring tasks shown in Appendix \ref{app: wolfalpharesults}. (`N-digit accurate' means that the answer differs from GT only by $\pm1$ on the $N-$th significant digit. For example, in Figure \ref{fig:Example of wolframalpha problem} Q4, the `w.o.wolfTool Answer' is 1-digit accurate, while the `w.wolfTool Answer' is 4-digit accurate, compared to the ground truth answer.)}
\label{tab: wolfalpharesults}
\end{table}

In all: experiments presented in this Section \ref{sec: impact study} show that, by integrating appropriate tools into the AI agent system, we can significantly advance the capabilities of LLMs in solving complex and challenging physics problems. This enhancement improves the mastery of physics of the agent system without requiring any modifications to the underlying LLMs. In conclusion, our Physics Supernova emerges as a powerful, flexible, and extensible physics problem-solving framework built upon the agent-based paradigm.

\section{Related Work}

\subsection{Task Solving Agents}%

Since LLMs were shown to be capable of performing a wide range of tasks with appropriate prompts~\cite{radford2019language}, researchers have been trying to enhance their capability and integration with tasks through prompt-based methods and tools. The concept of agent systems was then developed, exemplified by ReAct~\cite{yao2023react}, in which LLMs iterate between reasoning and tool calls to perform more complicated tasks~\cite{wang2025theoryagentstoolusedecisionmakers}. Later, agent-specific codebases have appeared: autogen~\cite{wu2023autogenenablingnextgenllm} provides framework for creating multi-agent systems; smolagents~\cite{smolagents} focuses more on code-related problem solving with defined domains. Upon these systems are more complicated general-purpose agent systems: for example, tool-creating agents like Alita~\cite{qiu2025alitageneralistagentenabling} focus on creating and using tools for complex tasks with self-improvement; products like Openai Deep Research~\cite{openai2025introducing} and manus~\cite{manus_2024} focus on providing reliable research results to users. There has also been work focusing on optimizing the cost efficiency and effectiveness of agent systems~\cite{actlessreasonmore,arpo,agentlightning}. These agent systems are mainly designed for general purpose tasks, with a focus on general virtual assistant related tasks represented by GAIA~\cite{mialon2023gaia}.

In addition to these general purpose agents, there is a growing body of work on domain-specific agents. For example, for math problems, some work has discussed the potential pipeline-based LLM approaches to mathematical problem solving~\cite{geminideepthinkimo2025,huang2025gemini25procapable,deepmind2024alphaproof}; for the humanities domain, HistAgent~\cite{histagent} introduces domain-specific high quality benchmarks and specially designed agent systems targeted at historical reasoning; for human-computer interaction and mental health, EmoAgent~\cite{emoagent} studies the impact of LLM-based chat systems on users' mental health, and proposes agent-based methods to monitor users' mental states and ensure safer human-AI interactions. These domain-specific agents show the potential of agent systems in solving professional tasks requiring specialized domain knowledge; however, previous work rarely focuses on solving expert physics problems with agent systems.

\subsection{Olympiads as Benchmarks}

Olympiad competitions (e.g. IMO, IPhO, etc.) are widely regarded as challenging even to domain experts, and have been seen as challenging benchmarks for LLMs. Recent work shows progress in Math Olympiad problems, with both natural-language-based pipelines~\cite{huang2025gemini25procapable} and formal-language-based provers~\cite{lin2025leanstarlearninginterleavethinking,chen2025seedproverdeepbroadreasoning}.

Physics Olympiads have gradually drawn more attention in recent years. OlympiadBench~\cite{he2024olympiadbenchchallengingbenchmarkpromoting} collects Olympiad problems across subjects, including physics. In 2025, SeePhysics~\cite{xiang2025seephysdoesseeinghelp} collects physics problems with images. Most benchmarks use publicly available problems from past Olympiads (e.g., prior IPhO problems). In contrast, PhyBench~\cite{qiu2025phybenchholisticevaluationphysical} curates an `entirely original' dataset, aiming to minimize data leakage; it excludes images. Some other datasets, represented by PhysicsArena~\cite{physicsarena} and PhysReason~\cite{physicsreason}, feature solutions with detailed reasoning processes. These benchmarks, though some contain problems that require image-related physics abilities, some require a high level of professional knowledge, and some require complicated reasoning and calculation, are primarily designed for nonagentic single-LLM settings.

\section{Discussions}

\subsection{Physics Experimental Exams: Instrument-based Exam and Program-based Exam as Proxies for Research-level Physics Experiments}

Experiments are foundational to physics research~\cite{newton1687principia}. In Physics Olympiads,\textbf{experimental exams} have served as proxies for research-level physics investigations. There are two common formats of Physics Olympiad experimental exams: \textbf{instrument-based Experiments} and \textbf{program-based Experiments}\footnote{Most Physics Olympiads use instrument-based experiments, for example, current IPhO, APhO, EuPhO, etc. During the pandemic, some Olympiads utilize program-based experiments, for example, APhO 2021, EuPhO 2020,  EuPhO 2021, etc.}. For the former, contestants are provided with (sometimes simplified) instruments and are asked to plan and carry out experiments and measurements; for the latter, contestants design and carry out simulated experiments with programs, followed by data analysis.

In this work, we mainly focus on the IPhO 2025 Theory Problems, rather than the instrument-based experimental problems. This limitation is caused in part by limited access to the experiment instruments. We hope with advances in robotics future LLM-based agents may also work on these experimental exams. Moreover, we argue that program-based exams could also serve as good proxies for benchmarking current and near-future AI's ability in research-level experiments. Although instrument-based experiments are closer to real-world research, there are \textbf{ two aspects} where program-based experiments are potentially better:

Program-based experimental exams can simulate more advanced and complex experiments compared to instrument-based exams. For instrument-based exams, contestants have to design experiments, conduct manual operations and measurements, process data, and answer questions.

For program-based experiments, contestants run simulated experiments through programs: Although manual experimentation is no longer required, critical challenges such as experimental design and data analysis continue to play a central role in program-based assessments. One big advantage is that program-based exams can access experiments that are more complex or require more sophisticated design because they are not limited by cost, safety concerns, or other limiting factors for instrument-based experiments. We provide two examples for instrument-based experiments and program-based experiments in Figure \ref{fig:Comparison between instrument-based and program-based experiments}, shown in Appendix \ref{app:Comparison between instrument-based and program-based experiments}.

Program-based experimental exams can shift the focus from testing robotic manipulation of instruments to evaluating the ability in "physics". There has been work using robotic systems to implement experiments in physics~\cite{uddin2025aidrivenroboticsfreespaceoptics}; however, in all, current AI systems still fail to perform robotic tasks to manipulate instruments in experiments. This critical challenge will be addressed with advances in robotics research. With program-based experiments, we hope that in the near future one can already better evaluate essential abilities of AI systems in experimental physics.

We also emphasize that while program-based experiments have potential advantages in benchmarking AI's capability in experimental physics, the instrument-based tests are certainly essential ultimately. As AI systems advance toward ASI and daily use, one should not overlook the importance of instrument-based tests as they
(1) yield a smaller deviation from real-world research; (2) provide a better metric for characterizing robotic capability; and (3) evaluate the performance better under extreme or unexpected conditions like instrument failure, etc.

\subsection{Verifiable Physics Reasoning}

In this work, we use an answer reviewer tool to verify the deductions of the worker, which is based purely on natural language. A huge step in automatic math proofs are \textbf{verifiable LLM-generated proofs} written in Lean~\cite{carneiro2024lean4leanverifiedtypecheckerlean}. Some previous work proposes to use Lean-like tools for verifiable physics formula deductions~\cite{song2025leancopilotlargelanguage,bobbin2023formalizingchemicalPhysicsusing}. However, the process of deriving physics formulas from natural-language-based problems, whether grounded in theoretical models or experimental observations, currently lacks reliable automatic verification through comparable processes. This limitation remains an open area for further research. Promising directions for future exploration include: (1) developing methods to verify the abstraction and transformation between formulas, physical representations, and intuitive reasoning; (2) establishing a more rigorous and transparent calculation framework that supports verifiability; and (3) enhancing answer-review systems with tools that possess broader and deeper expertise in physics.

The first direction represents a big challenge in symbolic AI ~\cite{symbolismai}. In relation to developing verifiable physics calculations, extensive work has been done on machine-checked mathematics in Lean-like languages~\cite{lean4}. For example, Deepseek-Prover, Kimina-Prover and AlphaProof~\cite{deepmind2024alphaproof,ren2025deepseekprov2,numina2025kimina} use RL-based methods to train LLMs that excel in generating lean-based proofs. Others discuss test-time scaling methods through pipeline workflows~\cite{zhou2025deltaprover,baba2025proveragent,ospanov2025apollo}. Future work for solving physics problems might adopt similar methods to improve the ability to generate reliable and verifiable solutions.

\textbf{In summary}, we suggest future work on using AI systems for physics problem-solving to focus on: (1) program-based or instrument-based experiments; and (2) verifiable and reliable solution generation.

\section{Conclusion}

In this work, we introduce Physics Supernova, a flexible agent system for solving Olympiad-level physics problems and beyond. By equipping the manager agent with task-specific tools including ImageAnalyzer, AnswerReviewer, etc., we extend the capability of state-of-the-art LLMs on physics problems. This aspect was previously viewed mostly as benchmarks of base, simple LLMs~\cite{xiang2025seephysdoesseeinghelp, he2024olympiadbenchchallengingbenchmarkpromoting, qiu2025phybenchholisticevaluationphysical}.

On the newly released IPhO 2025 theory problems, our agent system powered by Gemini 2.5 Pro achieves gold-medal-level performance. Specifically, our method ranks top $10\%$ among human contestants in all three theory problems. In total, our method achieves $23.5$ points on the theory problems, ranking $\#14$ among all $406$ human contestants, exceeding the median theory score of the gold medalists.

Our ablation study demonstrates the effectiveness of the tools provided. We further explore the possibility of improving Physics Supernova's ability on solving more complicated, knowledge-requiring problems by equipping it with more specialized tools, such as the WolframAlpha-based QA tool. The success shows that the agent-oriented paradigm offers a \textbf{powerful and flexible} platform for the integration of tools for the solution of advanced physics problems.

Overall, our proposed Physics Supernova successfully improves the physics problem solving ability of LLMs through agent paradigms. This shows the potential of agent systems to improve the capability of LLMs for scientific reasoning and physics-related tasks, and further implies their potential for developing super intelligence that embeds into the real world.

\newpage
\bibliographystyle{unsrt}
\bibliography{reference.bib}

\newpage
\appendix

\section{Examples of IPhO 2025 Problems Scoring Criteria}
\label{app: score criteria example}

\begin{figure}[h]
  \centering
  \includegraphics[width=1.0\linewidth]{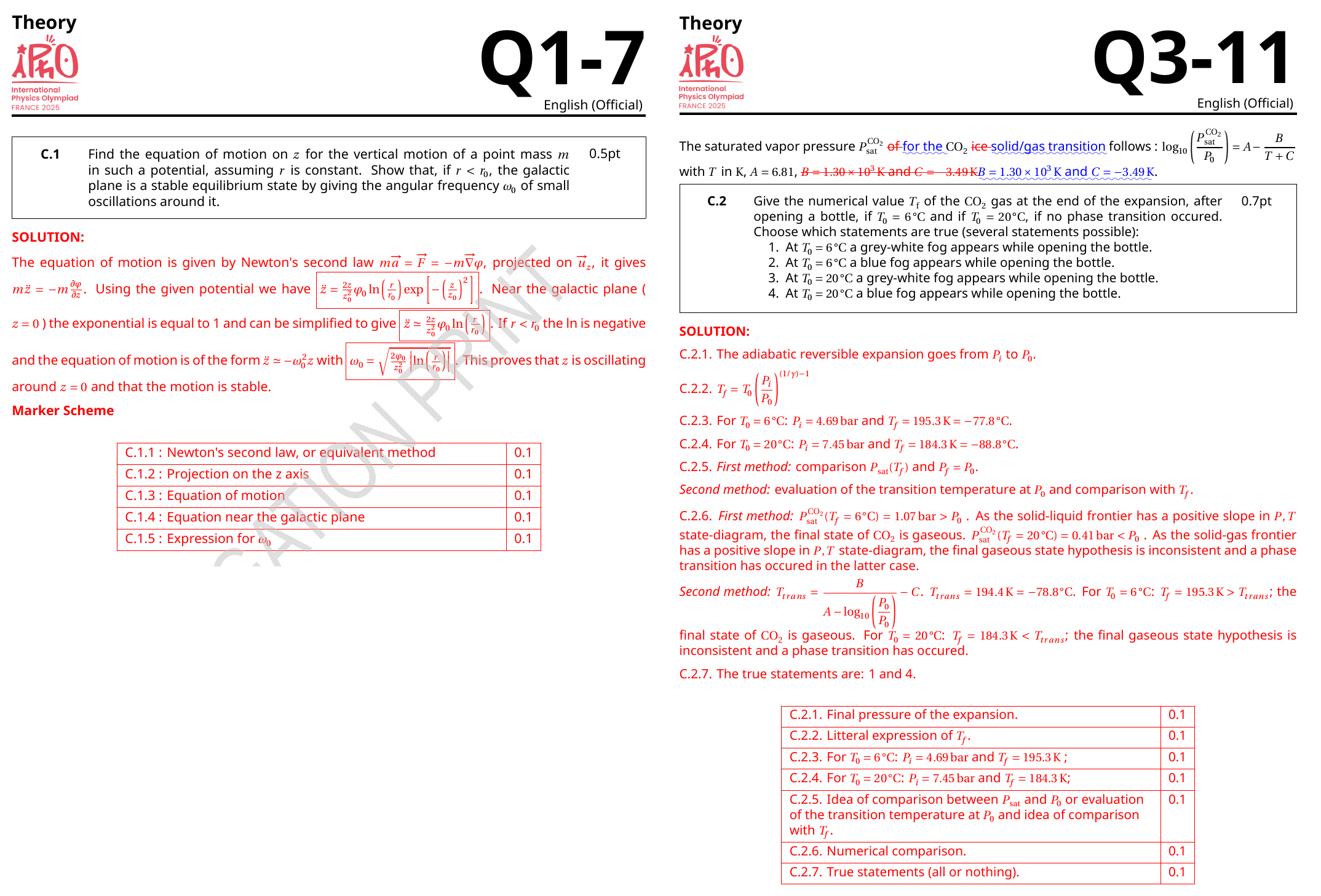}
  \caption{Two scoring criteria examples, for IPhO 2025 Theory Problem 1 Part C.1 and Theory Problem 3 Part C.2, correspondingly. As shown, there are $5$ scoring criteria for Theory Problem 1 Part C.1, and $7$ scoring criteria for Theory Problem 3 Part C.2.}
  \label{fig: scoring criteria example1}
\end{figure}

We provide two examples of the IPhO 2025 scoring criteria (corresponding to Theory Problem 1 Part C.1 and Theory Problem 3 Part C.2, respectively) in Figure \ref{fig: scoring criteria example1}, obtained from \url{https://ipho.olimpicos.net/}.

As shown in Figure \ref{fig: scoring criteria example1}, for each problem, very detailed answers and scoring criteria are provided, making fine-grained scoring possible for answers. In Table \ref{tab:theory problem description} we count the number of scoring criteria for each part of each Theory Problem.

\section{Detailed Prompts}
\subsection{Image Analyzer Tool Prompt}
\label{app: image analyzer tool prompt}
The image analyzer tool utilizes an LLM provided with the image and the question from the manager agent. Its task is to answer the manager's questions based on the provided information. It then returns an str-object of its measurements.

\begin{python}
# input: manager_query: str, img_file: ImageFile, vision_expert_llm: LLMModel
IMG_SYSTEM_PROMPT = "You are an expert in dealing with image in Physics Olympiads."
messages = [
    ChatMessage(role=MessageRole.SYSTEM, content=IMG_SYSTEM_PROMPT),
    ChatMessage(role=MessageRole.USER, content=[
        {"type": "image", "image": img_file},
        {"type": "text", "text": question},
    ]),
]
output: str = vision_expert_llm.generate(messages)
\end{python}

\subsection{Answer Reviewer Tool Prompt}
\label{app: answer reviewer tool prompt}
The Answer Reviewer tool utilizes an LLM provided with: (1) manager agent's solution; (2) manager agent's notes; (3) the original problems (including texts and figures). It then returns an str-object representing its review results.
\begin{python}
# input: agent_solution:str, agent_note:str, markdown_content: List[Dict[str, Any], review_expert_llm: LLMModel
# markdown_content includes markdown file text and image.
REVIEW_SYSTEM_PROMPT = (
    "You are an uncompromising Physics peer-reviewer. Your job is to find *every* logical, mathematical 
error in the worker's answer. "
    "Check dimensional consistency, missing steps, incorrect sign conventions, numerical mistakes, and 
unclear explanations. Focus especially on wrong answers, less on presentations."
    "Be extremely critical: if something is wrong, point it out and request clarification or correction. 
Mainly focus on errors that would lead to a wrong result, rather than focusing extremely on presentation 
or style."
    "It is possible that the worker's answer is not correct, so please be prepared to provide detailed 
feedback. The worker's answer contains some error, so you must check and point it out. Also, if the 
worker reads measurements from image, make sure to remind the worker that whenever it reads or measures 
from image, it uses the ask_image_expert tool, or the readings might be very inaccurate.\n"
)

review_instruction = (
    f"Please review the following solution:\n\n"
    f"WORKER'S SOLUTION:\n{agent_solution}\n\n"
    f"WORKER'S NOTE: {agent_note}\n\n"
    f"Please provide detailed feedback on correctness. "
    f"Point out any errors, wrong steps, focus more on correctness rather than presentation."
    f"The original problem follows:"
)

combined_content : List[Dict[str, Any]] = [
    {"type": "text", "text": review_instruction}
] + markdown_content

messages = [
    ChatMessage(role=MessageRole.SYSTEM, content=REVIEW_SYSTEM_PROMPT),
    ChatMessage(role=MessageRole.USER, content=combined_content),
]
output: str = review_expert_llm.generate(messages)
\end{python}

\section{Expert-knowledge Requiring Tasks}
\label{app: wolfalpharesults}

We further generate several tasks requiring expert knowledge to test how integrating WolframAlpha would help Physics Supernova with accurate expert knowledge.

As shown in the following examples, \textbf{When given access to WolframAlpha Tools, the agent system provides more accurate answers.} In the experiments, we use Gemini 2.5 Pro as LLM and compare the result with and without WolframAlpha Tools, as shown below. The aggregated results are also shown in Table \ref{tab: wolfalpharesults}.

\begin{qaBlock}
\textbf{\problemname~Q1.}

Using the latest AME (Atomic Mass Evaluation) atomic masses, compute the Q-value of double beta decay $^{76}Ge \rightarrow ^{76}Se + 2 e^-$ (ground state → ground state).
Return a single number: the value in keV, rounded to exactly 5 significant digits, in scientific notation. Do not include units or extra text.

\vspace{2pt}\hrule height 0.4pt \vspace{2pt}
\AnswerAligned{:}{$2.0391$E+3}{$2.0391$E+3}{$2.0390$E+3}
\end{qaBlock}

\begin{qaBlock}
\textbf{\problemname~Q2.}

Using NIST XCOM (or an equivalent authoritative database), determine the mass attenuation coefficient $\mu/\rho$ of lead (Pb) for photons of energy 662.0 keV ($^{137}Cs$ $\gamma$ line).
Return a single number: the value in $cm^2·g^{-1}$, rounded to exactly 5 significant digits, in scientific notation. Do not include units or extra text.

\vspace{2pt}\hrule height 0.4pt \vspace{2pt}
\AnswerAligned{:}{$1.1105$E-1}{$1.1352$E-1}{$1.1150$E-1}
\end{qaBlock}

\begin{qaBlock}
\textbf{\problemname~Q3.}

Using the Ciddor (1996) refractive-index model for air, at wavelength $\lambda = 633 nm$ (vacuum), P = 101325 Pa, T = 20 °C, RH = 50 , and $\text{CO}_2$ = 450 ppm, compute $n-1$.
Return a single number: the value (dimensionless), rounded to exactly 5 significant digits, in scientific notation. Do not include units or extra text.

\vspace{2pt}\hrule height 0.4pt \vspace{2pt}
\AnswerAligned{:}{$2.7132$E-4}{$2.6894$E-4}{$2.7139$E-4}
\end{qaBlock}

\begin{qaBlock}
\textbf{\problemname~Q4.}

Using the IAPWS-IF97 formulation for water/steam, compute the specific enthalpy of water at p = 15 MPa and T = 650 K (single-phase state as appropriate).
Return a single number: the value in $kJ·kg^{-1}$, rounded to exactly 5 significant digits, in scientific notation. Do not include units or extra text.

\vspace{2pt}\hrule height 0.4pt \vspace{2pt}
\AnswerAligned{:}{$2.8686$E+3}{$3.0462$E+3}{$2.8690$E+3}
\end{qaBlock}

\begin{qaBlock}
\textbf{\problemname~Q5.}

Using NIST X-ray transition energies (or equivalent), determine the photon energy of the copper $\kappa\alpha_1\ (KL_3)$ line for elemental Cu at ambient conditions.
Return a single number: the value in keV, rounded to exactly 5 significant digits, in scientific notation. Do not include units or extra text.

\vspace{2pt}\hrule height 0.4pt \vspace{2pt}
\AnswerAligned{:}{$8.0478$E+0}{$8.0463$E+0}{$8.0478$E+0}
\end{qaBlock}

\begin{qaBlock}
\textbf{\problemname~Q6.}

Using the IGRF 13th generation (epoch 2025.0), compute the total geomagnetic field magnitude at (40.0140° N, 105.2705° W, altitude 1624 m) on 2025-01-01 00:00 UTC.
Return a single number: the value in nT, rounded to exactly 5 significant digits, in scientific notation. Do not include units or extra text.

\vspace{2pt}\hrule height 0.4pt \vspace{2pt}
\AnswerAligned{:}{$5.1321$E+4}{$4.9726$E+4}{$5.1300$E+4}
\end{qaBlock}

\begin{qaBlock}
\textbf{\problemname~Q7.}

Using CODATA-2022 fundamental constants, compute the rest frequency of the neutral hydrogen 21 cm hyperfine transition (ground-state spin-flip).
Return a single number: the value in Hz, rounded to exactly 5 significant digits, in scientific notation. Do not include units or extra text.

\vspace{2pt}\hrule height 0.4pt \vspace{2pt}
\AnswerAligned{:}{$1.4204$E+9}{$1.4228$E+9}{$1.4204$E+9}
\end{qaBlock}

\begin{qaBlock}
\textbf{\problemname~Q8.}

Using the NIST ESTAR database (or equivalent), determine the mass stopping power of aluminum (Al) for electrons of kinetic energy 1.000 MeV.
Return a single number: the value in $MeV·cm^2·g^{-1}$, rounded to exactly 5 significant digits, in scientific notation. Do not include units or extra text.

\vspace{2pt}\hrule height 0.4pt \vspace{2pt}
\AnswerAligned{:}{$1.4860$E+0}{$1.5980$E+0}{$1.5980$E+0}
\end{qaBlock}

\begin{qaBlock}
\textbf{\problemname~Q9.}
Using JANAF/NIST thermochemical data (ideal-gas heat capacities), determine the molar heat capacity at constant pressure, $C_p$, of nitrogen gas ($N_2$) at T = 1200 K (assume thermally perfect ideal gas, no dissociation).
Return a single number: the value in $J·mol^{-1}·K^{-1}$, rounded to exactly 5 significant digits, in scientific notation. Do not include units or extra text.
\vspace{2pt}\hrule height 0.4pt \vspace{2pt}
\AnswerAligned{:}{$3.3723$E+1}{$3.3540$E+1}{$3.3724$E+1}
\end{qaBlock}

\begin{qaBlock}
\textbf{\problemname~Q10.}

For Beijing, China (39.9042° N, 116.4074° E, elevation 50 m), determine the umbral magnitude of the next lunar eclipse after 2025-08-09 that is at least partially visible from that location.
Return a single number: the umbral magnitude (dimensionless), rounded to exactly 5 significant digits, in scientific notation. Do not include units or extra text.

\vspace{2pt}\hrule height 0.4pt \vspace{2pt}
\AnswerAligned{:}{$1.3638$E+0}{$1.1510$E+0}{$1.3680$E+0}
\end{qaBlock}

\begin{figure}[!h]
  \centering
  \includegraphics[width=0.8\linewidth]{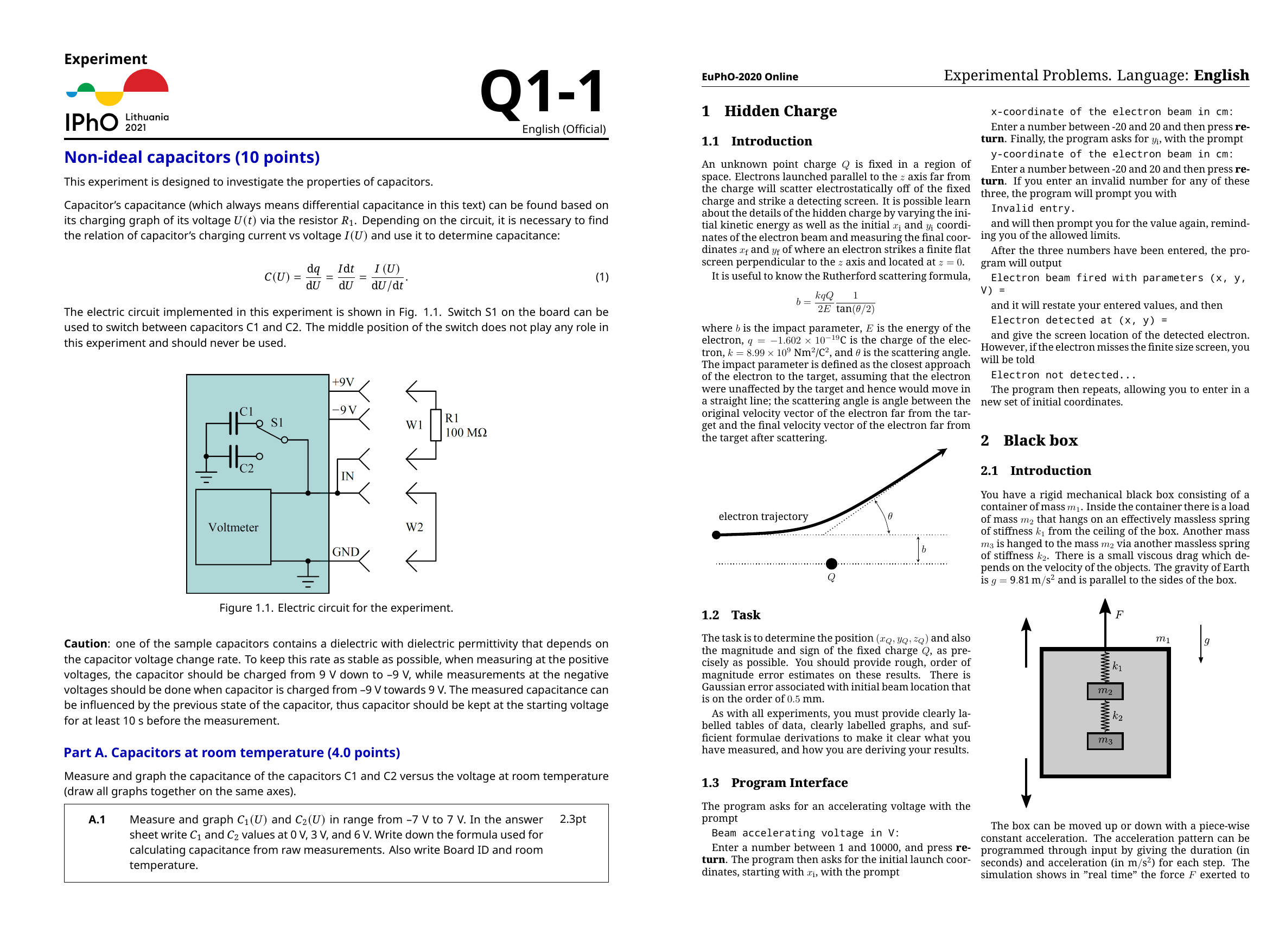}
  \caption{Left: an instrument-based experiment example (IPhO 2021 experiment problem 1); Right: a program-based experiment example (EuPhO 2020 experiment problem 1). For IPhO 2021 experiment problem 1, a circuit board with electronic components to be measured inside it is provided, where contestants have to conduct measurements for these components: \textbf{this instrument-based experiment requires contestants to appropriately conduct manipulations on real experiment instruments, which is not tested in program-based experiments.} For EuPhO 2020 experiment problem 1, a program simulates experiments about detecting unknown charge with electron beams, similar to the Rutherford scattering experiment: \textbf{this program-simulated experiment is more related to modern physics, and it is impractical in a typical Olympiad venue due to cost and safety constraints.}}
  \label{fig:Comparison between instrument-based and program-based experiments}
\end{figure}

\section{Example of Instrument-based and Program-based experiments}
\label{app:Comparison between instrument-based and program-based experiments}

We provide two examples of Instrument-based and Program-based experiments, as shown in Figure \ref{fig:Comparison between instrument-based and program-based experiments}: the instrument-based experiment is IPhO 2021 experiment problem 1; and the program-based experiment is EuPhO 2020 experiment problem 1.

As shown and described in the caption of Figure \ref{fig:Comparison between instrument-based and program-based experiments}, the program-based experiment can be more related to modern physics and bypasses the difficulties of cost and safety issues, although they are less real compared to instrument-based experiments.

\end{document}